\documentclass[conference]{IEEEtran}

\ifCLASSINFOpdf

\else

\fi

\usepackage[cmex10]{amsmath}
\usepackage{amsthm}
\usepackage{amssymb}
\usepackage{mathrsfs}
\usepackage[dvips]{graphicx}
\usepackage{float}
\usepackage{array}
\usepackage[T1]{fontenc}
\usepackage{lipsum}  %for table
\usepackage{placeins}
\usepackage{multirow}

\begin{document}
%\lipsum[1]
%\linespread{1.8}\selectfont
\title{A Geometric Approach To Fully Automatic Chromosome Segmentation}

\author{\IEEEauthorblockN{Shervin Minaee}
\IEEEauthorblockA{ECE Department\\ 
New York University\\Brooklyn, New York, USA\\
shervin.minaee@nyu.edu\\}
\and
\IEEEauthorblockN{Mehran Fotouhi}
\IEEEauthorblockA{Computer Engineering Department 
\\Sharif University of Technology\\Tehran, Iran\\
mehran.fotouhi@gmail.com\\}
\and
\IEEEauthorblockN{Babak Hossein Khalaj}
\IEEEauthorblockA{Electrical Engineering Department 
\\Sharif University of Technology\\Tehran, Iran\\
khalaj@sharif.edu\\}
}

\maketitle

\begin{abstract}
A fundamental task in human chromosome analysis is chromosome segmentation. Segmentation plays an important role in chromosome karyotyping. The first step in segmentation is to remove intrusive objects such as stain debris and other noises. The next step is detection of touching and overlapping chromosomes, and the final step is separation of such chromosomes. Common methods for separation between touching chromosomes are interactive and require human intervention for correct separation between touching and overlapping chromosomes.
In this paper, a geometric-based method is used for automatic detection of touching and overlapping chromosomes and separating them. The proposed scheme performs segmentation in two phases. In the first phase, chromosome clusters are detected using three geometric criteria, and in the second phase, chromosome clusters are separated using a cut-line. 
Most of earlier methods did not work properly in case of chromosome clusters that contained more than two chromosomes. Our method, on the other hand, is quite efficient in separation of such chromosome clusters. At each step, one separation will be performed and this algorithm is repeated until all individual chromosomes are separated. Another important point about the proposed method is that it uses the geometric features of chromosomes which are independent of the type of images and it can easily be applied to any type of images such as binary images and does not require multispectral images as well. We have applied our method to a database containing 62 touching and partially overlapping chromosomes and a success rate of 91.9\% is achieved.
\end{abstract}

% \fi
\IEEEpeerreviewmaketitle

\section{Introduction}
\IEEEPARstart{C}{hromosome} karyotyping is an essential task in cytogenetics and is usually performed in clinical and cancer cytogenetic labs and can be used in the diagnosis of genetic disorders. The normal human karyotypes contain 22 pairs of autosomal chromosomes and one pair of sex chromosomes. Chromosome karyotyping is meant to identify and assign each chromosome in the image to one of the 24 classes.
Chromosome karyotyping has three main steps: pre-processing, segmentation and classification. Among these steps, chromosome segmentation is very  important, since it affects performance of classification which is the final goal. Chromosome images may have some defects; they may be bent, they may touch or overlap and their bands may be spread. In addition, since touching and overlapping chromosomes exist in almost every metaphase image, the solution of this problem is vital. The first step in analyzing a chromosome image is  segmentation of chromosomes from the image background, the main methods used in this step are based on the evaluation of a global threshold by means of the Otsu method \cite{1}, or on a re-thresholding scheme \cite{2}. Due to the fact that long chromosomes may touch and overlap, the first segmentation step is usually unable to identify each chromosome as a single object, and  presents a number of clusters. So far, attempts have been made to deal with clusters of touching (but not overlapping) chromosomes  \cite{3}, \cite{4}, \cite{5}, and for clusters of overlapping (but not touching) chromosomes \cite{6}, \cite{7}, where both of geometric and intensity based features have been used to resolve segmentation ambiguities. Lerner \cite{8} proposed a method to combine the choice of correct cluster disentanglement with the classification stage, resulting in a classification-driven segmentation. Grisan \cite{9} proposed a similar method. There are many other methods for separation between touching and overlapping objects \cite{10}, \cite{11}. Schwartzkopf \cite{12} proposed a method for joint segmentation and classification that used statistical method. Since this method was applied to multispectral chromosome images, it does not work for binary images. So far,  most of chromosome analysis systems have a common fault: their poor automatic chromosome incision ability. Most of current systems for automatic chromosome segmentation are interactive and need human intervention.
We have to mention that the original images are pre-processed and the chromosomes are segmented from the background and the intrusive objects and noises are removed from the background. Therefore, our main effort is to detect and separate touching or overlapping chromosomes.

It is worth mentioning that there are different approaches for segmentation and classification of medical images. One main approach is to the geometric characteristics of the object of interest, the other one is using spatial and transform domain features of the image, etc. The right set of features depends on the application. For example, in biometric recognition area, there are a lot of works based on spatial and frequency domain information of images. One such work is presented in \cite{13}, where the author uses the spatial and wavelet domain features of images to perform palmprint recognition. However, some of those approaches requires a very large dataset to train the model properly so they may not be applicable to small datasets, because they could be very prone to over-fitting. A good work for dealing with small dataset is presented in \cite{14}, where the author explains how to jointly maximize the model accuracy and reliability .
In this paper, a geometric method for segmentation of the touching and partially overlapping chromosomes is presented. First, we introduce an approach to evaluate whether an object is a single chromosome or a chromosome cluster. By chromosome clusters, we mean a group of chromosomes which overlap and touch each other. Subsequently, for each cluster, we use geometric features of chromosome boundary which help separate touching or partially overlapping chromosomes. Chromosome segmentation is performed in two phases. In the first phase, touching or overlapping chromosomes are detected using the approach which is introduced in Section II where we deal with the chromosomes' shape and their geometric features. If two or more chromosomes overlap, the resulting cluster would not have the usual long and thin shape and we can use such difference to detect chromosome clusters. In the second phase, we use other geometric features to separate touching or partially overlapping chromosomes. We will discuss about this step in Section III which deals with boundary pattern of chromosomes.\\
Our method has three advantages over earlier schemes:
\begin{enumerate}
\item	First, it can be applied to any type of images, even binary image, and it does not need multispectral or grayscale images. Therefore it can reduce the cost of photography and the amount of computation.
\item	Second, it can easily separate chromosome clusters that contain more than two chromosomes where most earlier schemes fail.
\item Third, our method is fully automatic and does not need any human intervention.
\end{enumerate}

\section{Detection of touching or overlapping chromosomes}
In order to detect chromosome clusters, we use three criteria which deal with the geometry of chromosomes. The first method is surrounding ellipse method (Section II.A), which is based on the ratio of the length of minor axis of surrounding ellipse to the length of its major axis. The second method is convex hull method (Section II.B) which is based on the number of pixels in the original chromosome to the number of pixels in its convex hull ratio. An important point about this method is its robustness in detecting small chromosomes that may produce error in the first method. The third method is skeleton and end points (Section II.C), which uses the skeleton of each chromosome (either single chromosome or a chromosome cluster) to find the end points of skeleton and decides based on the number of end points.
All of these methods have some limitations, but through proper integration, we can detect all chromosome clusters (either touching or overlapping chromosomes) as shown through our simulation results. 
Each chromosome passes through these three methods and in case it satisfies the criteria of all three methods, it will be detected as a chromosome cluster. We will discuss the details of each method in the following parts.

\subsection{Surrounding ellipse method}
Surrounding ellipse of a shape is an ellipse which surrounds that. Single chromosome is usually long and thin (unless those chromosomes which belong to 20th, 21st or 22nd group) so its surrounding ellipse will be long, but the overlapping chromosomes have a surrounding ellipse close to a circle. We can use this difference for detection of overlapping chromosomes. In order to take advantage of such difference, the ratio of the length of minor to major axis of the surrounding ellipse has to be found. If the label is overlapping, we expect this ratio to be close to 1, because the surrounding ellipse would be close to a circle, but if the chromosome is single it will have a smaller ratio. Therefore, a threshold can be determined to distinguish between chromosome clusters and single chromosomes.
We propose the below algorithm for this step:
\begin{enumerate}
\item	Find the surrounding ellipse of each chromosome label

\item	Find the ratio of minor axis length to major axis length of surrounding ellipse for each chromosome (either a single chromosome or chromosome cluster)

\item	Determine a threshold (we simply set this threshold to the average of all ratios, but if we have a large dataset, we can use a training set to determine this threshold)

\item Compare ratio of each label with this threshold. For each label, if the ratio is less than the threshold, remove it, but if the ratio is more than the threshold, keep this chromosome
\end{enumerate}
This method is very fast, but it has problems with two types of chromosomes:
\begin{enumerate}
\item	Small chromosomes

\item	Bent chromosomes
\end{enumerate}

The shape of small chromosomes is different from usual chromosomes as they have a round shape where the ratio of minor axis length to major axis length of their surrounding ellipse will be similar to overlapping chromosomes.
Bent chromosomes also have ratios similar to overlapping chromosomes so they may wrongly be detected as overlapping chromosomes, an issue that needs to be addressed properly. As the proposed algorithms are applied to each chromosome in a cascade fashion, each step has to remove those single chromosomes which are not removed in the previous steps.

\subsection{Convex hull method}

In Euclidean space, an object is convex if for every pair of points within the object, every point on the straight line segment that joins them is also within the object.
The convex hull of a set C is the smallest convex set that contains C. Convex hull have been used in several applications in computer vision, image analysis, and digital image processing, including object recognition, image and video coding. %\cite{16}% \cite{15}
As a normal chromosome has a relatively convex shape, its convex hull would approximately have the same number of pixels as the original chromosome. If we find the convex hull of the chromosomes, we will notice that the convex hull of chromosome clusters have much more pixels than chromosome clusters themselves, whereas the single chromosomes have almost the same number of pixels as their convex hulls. Consequently, we can detect chromosome clusters using such difference. In order to achieve this goal, we should find the ratio of the number of pixels in each chromosome to number of pixel in its convex hull for all chromosomes and then compare these ratios with a threshold. For each chromosome, if the ratio is less than a given threshold, we expect that this label would be an overlapping chromosomes and vice versa. The proposed algorithm for this method is given below:
\begin{enumerate}
\item	Find convex hull of each chromosome label.

\item	Calculate the ratio of the number of pixels in the original chromosome to the number of pixel in its convex hull for each chromosome.

\item	Determine a threshold (this threshold can be determined using training set, or it can simply set to the average of these ratios for all chromosomes)

\item Compare this ratio for each chromosome with this threshold, for each label if the ratio was more than the threshold eliminate this chromosomes.

\item	The remaining chromosomes will be sent to the next step.
\end{enumerate}

One advantage of this method is that we can eliminate small single chromosomes remaining from the previous step. Since for these chromosomes the convex hull is almost coincident with original chromosome, the ratio will be more than the given threshold. However, as we still have problem with bent chromosomes, we should eliminate them in the next step.
Fig.1 represents an image of chromosomes with convex hulls of two chromosomes.

\begin{figure}[1 h]
\begin{flushleft}
    \includegraphics [scale=0.48] {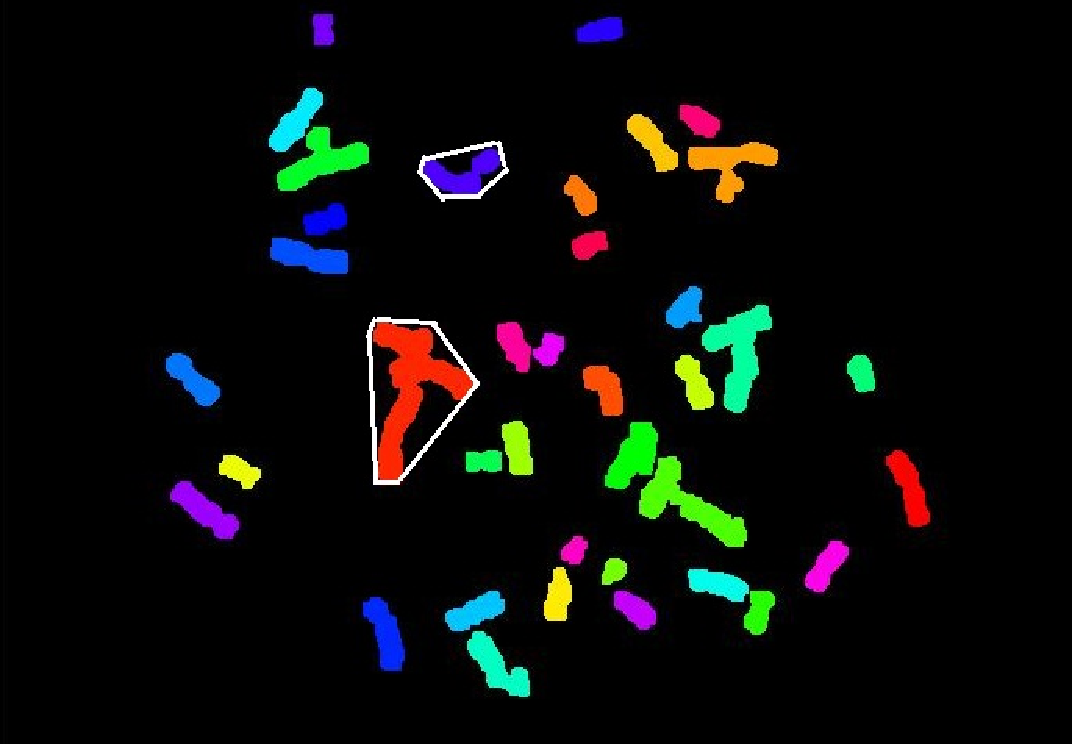}
 \end{flushleft}
  \caption{A chromosome image with convex hull of two chromosomes}
\end{figure}

\subsection{Skeleton and end points}
Skeletonization is the transformation of a component in a digital image into a subset of the original component. Skeleton has been used in several applications in computer vision, image analysis, and digital image processing.

We used this method as one step of the chromosome clusters detection algorithm. If we find the skeleton of each chromosome and then find the end points of this skeleton (end points are those point which are the last point in any side of a line) we will notice that the overlapping chromosomes usually have more than 2 end points. Therefore, we can use this idea for detection of  overlapping chromosomes. Skeletons and end points of a set of chromosomes are represented in Fig.2. End points of chromosomes are shown with red points. We observe that all chromosomes clusters in this picture have more than two end points.
\begin{figure}[2 h]
\begin{flushleft}
    \includegraphics [scale=0.32] {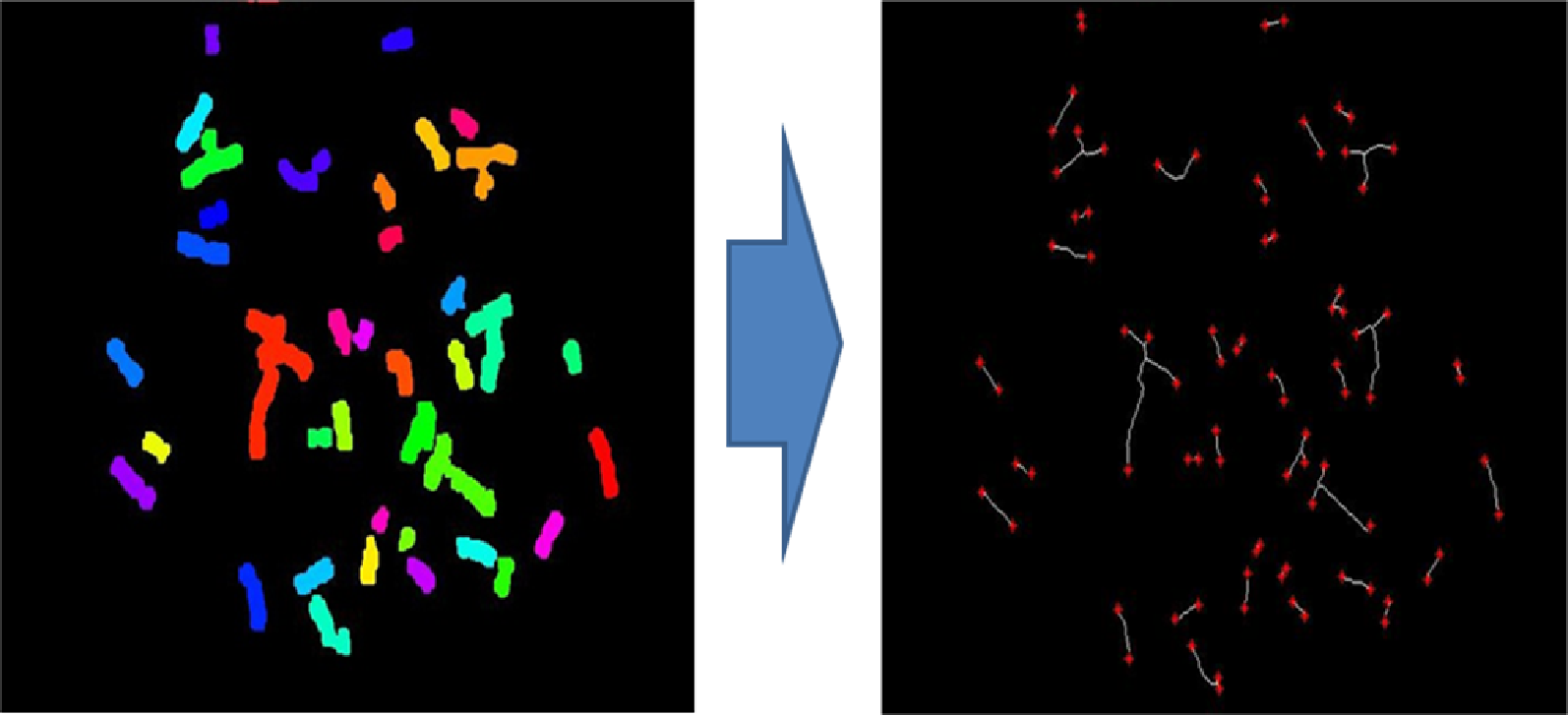}
 \end{flushleft}
  \caption{An example of skeletons and end points of a chromosome image}
\end{figure}

The proposed algorithm for this method is:
\begin{enumerate}
\item	Find the skeleton of each chromosomes.

\item	Find the end points of each skeleton.

\item	In the case more than two points are found, classify them as a chromosome clusters.
\end{enumerate}

This method is robust for finding overlapping chromosomes. However, because of the iterative structure of the skeleton algorithm, it is time-consuming and we should improve its computational complexity. In the next part, we combine these three methods in a proper way.

\subsection{Integration Step}

%All previously stated methods have their own problems. Surrounding ellipse method has problems with small or bent chromosomes, convex hull method has problem with bent chromosomes and skeleton method is too time-consuming.%
In order to solve the problem of the skeleton method, we decided to apply this method to a fewer number of chromosomes, initially 40 to 46 chromosomes, some of which are overlapping. First, we apply convex hull and surrounding  ellipse methods and eliminate a large number of single chromosomes. After these two steps we usually have about 8 to 14 chromosomes. Subsequently, the skeleton method can be applied to detect chromosome clusters from the remaining ones and because the number of chromosomes has been reduced from 46 to between 8 and 14, we will improve the time efficiency of the algorithm by a factor of 4. %Therefore, we will apply skeleton method after applying two other methods.
On the other hand, the surrounding ellipse method can not eliminate small single chromosomes and it is better to apply surrounding ellipse method after convex hull method. The block diagram for the direction of overlapping chromosomes shown in Fig.3 :

\begin{figure}[3 h]
\begin{flushleft}
    \includegraphics [scale=0.45] {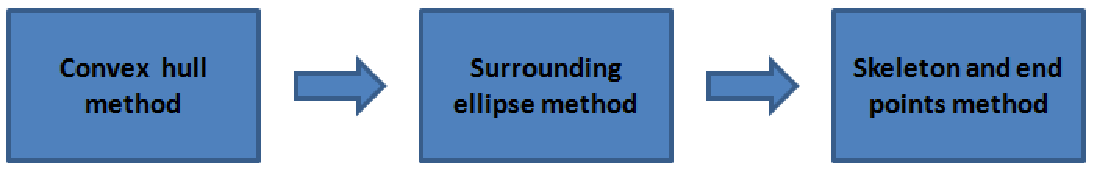}
 \end{flushleft}
  \caption{Block diagram of chromosome clusters detection algorithm}
\end{figure}

Each chromosome passes through these three methods and if it satisfies all three criteria, it will be considered as a chromosome cluster. After detection of all chromosome clusters, they will be used as the input of the second phase, which is separation of chromosome clusters.

\section{Separation of touching or partially overlapping chromosomes using cut-line method}
As discussed previously, after detection of overlapping chromosomes we need to separate them.

We introduce another geometric-based method for separation of touching or partially overlapping chromosomes. First, we find the cross-points of overlapping chromosomes. Cross-points are those points on the boundary of a chromosome cluster where two chromosomes touch or overlap. We will discuss the methods which can be used to find these cross-points in Sections III.A and III.B. Through the application of the proposed method, various touching or partially overlapping chromosomes can be handled in the same way. Two chromosome clusters and their cross-points are shown in Fig.4:

\begin{figure}[5 h]
\centering
    \includegraphics [scale=0.8] {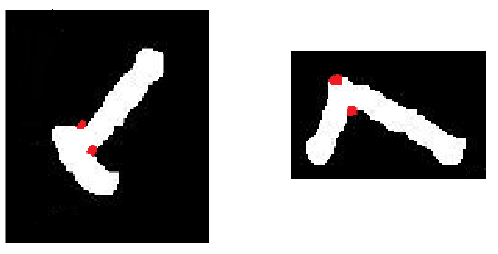}
  \caption{Cross-points of chromosome clusters}
\end{figure}

Once the cross-points are found, we should separate chromosome clusters using these cross-points. If the chromosome cluster consists of two chromosomes, it can then be cut from the line between cross-points resulting in two single chromosomes. However, if it consists of more than two chromosomes, we should repeat the whole algorithm multiple times.

In the following sections, we will introduce some approaches to find such cross-points. In order to find these cross-points, we only need to search on the boundary of the chromosomes. Therefore, in order to reduce the amount of computations, we can extract the boundary of chromosomes and search for the cross-points only in the boundary locations. Once the boundary extraction is done, we can sort the pixels on the boundary in a clockwise fashion. Suppose that the sorted boundary pixels are located in an N$\times$2 matrix in which each row contains the coordinates of the i-th pixel on the boundary and N is the total number of pixels on the boundary. This boundary matrix will be denoted by B. Fig.5 illustrates the result of boundary extraction in a chromosome cluster.

\begin{figure}[6 h]
\begin{center}
    \includegraphics [scale=0.55] {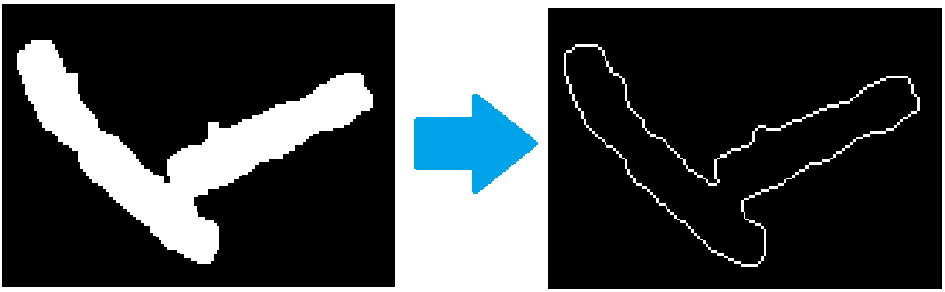}
 \end{center}
  \caption{A chromosome cluster and its boundary}
\end{figure}

In order to find the cross-points, we use two criteria based on the geometry of the boundary. The criteria are:
\begin{enumerate}
\item	Variations in the Angle of Motion Direction (VAMD)
%\item Non-Convexity of Chromosome (NCC)
\item	Sum of Distances among Total Points (SDTP)
\end{enumerate}
The first criterion, VAMD, is explained in Section III.A. It tries to find the cross-points based on the variation in the angle of motion. The second criterion is SDTP as explained in Section III.B. This criterion uses the fact that cross-points are usually located in the middle of a chromosome cluster. All pixels of the boundary pass through these two criteria, and at each step, some of boundary points will be eliminated and the total cross-points will be selected with a cost function which takes into account both these criteria. %among the remaining points. 

\subsection{Variations in the Angle of Motion Direction (VAMD)}

In order to understand the meaning of VAMD, suppose that an object is moving on the boundary of a chromosome. At each pixel, it has to move in a direction called motion direction which leads it to the next pixel. The angle between this direction and the horizontal axis is called the angle of motion direction. This angle can be calculated as the angle of connecting line between i-th and (i+1)-th pixels. We denote this angle with $\theta_{i}$.

\begin{align}
\theta_{i} = \tan^{-1}{\left(\frac{y(i+1) - y(i)}{x(i+1) - x(i)} \right)}
\end{align}

It should be noted that due to noise on the boundary, it is better to use more pixels to find a better estimation of the angle of motion direction. One can use angle of the connecting line between i-th pixel and j-th pixel as:

\begin{align}
\theta_{i}^{(j)} = \tan^{-1}{\left(\frac{y(j) - y(i)}{x(j) - x(i)} \right)}
\end{align}

Subsequently, we can use a weighted average of $\theta_{i}^{(j)}$ for different j's to find a robust estimation of $\theta_{i}$.

\begin{align}
\theta_{i}= \frac{\sum_{j=i+1}^{i+N_1} \beta_{i}^{(j)} \theta_{i}^{(j)}+\sum_{j=i-N_2}^{i-1} \alpha_{i}^{(j)} \theta_{i}^{(j)}}{\sum_{j=i+1}^{i+N_1} \beta_{i}^{(j)}+\sum_{j=i-N_2}^{i-1} \alpha_{i}^{(j)}}
\end{align}

The first summation is the estimated $\theta_{i}^{(j)}$ using the forthcoming pixels and the second summation is the estimated $\theta_{i}^{(j)}$ using previous pixels. The weights $\beta_{i}^{(j)}$ and $\alpha_{i}^{(j)}$ can be set to a fixed value or can be adaptive. The adaptive choice usually works better and it has to be a function of the Euclidean distance between the i-th and j-th pixels. For example, one possible choice of $\beta_{i}^{(j)}$ and $\alpha_{i}^{(j)}$ could be $e^{-(d(B_i,B_j))^2}$,
where $d(B_i, B_j)$ denotes the Euclidean distance between i-th and j-th pixels. As can be verified from the above formula, for pixels with long distance from the current pixel, $d(B_i, B_j)$ would be large, therefore the corresponding weights would be very small, which is reasonable.

Based on our simulation we deduced that if j=i+4, i+5 are used, the result will be the most satisfactory. Therefore, we used the following formula to find $\theta_{i}$:
\begin{align}
\theta_{i}=(1/2)\times( \theta_{i}^{(i+4)}+\theta_{i}^{(i+5)} )
\end{align}

Fig.6 depicts this method on a curve.
\begin{figure}[7 h]
\begin{center}
    \includegraphics [scale=0.55] {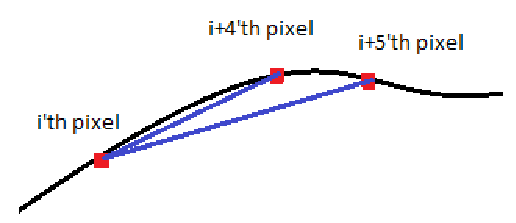}
 \end{center}
  \caption{Representation of boundary pixel's angle}
\end{figure}

After finding $\theta_{i}$ for all pixels on the boundary, we have to calculate the variation of angle in the i-th pixel as the difference of the (i+1)-th pixel angle and the i-th pixel.
\begin{align}
\Delta\theta_{i} \ = \theta_{i+1} \  - \theta_{i}
\end{align}

We expect to have a larger $\Delta\theta_{i}$ in cross-points compared to the other points of the boundary.
We can use the following algorithm to remove superfluous points on the boundary:
\begin{enumerate}
\item	Find the variation of angle in each pixel on the boundary.

\item Calculate the average of $\Delta\theta : \quad \Delta\theta_{avg} = \frac{\sum_{i=1}^N \Delta\theta_{i}}{N } $.

\item	For each pixel if  $\Delta\theta_{i} < \lambda_1 \Delta\theta_{avg} $ 
then remove this pixel from candidate pixels for cross-points (we used $\lambda_1$=1 which is found by trial and error).
\end{enumerate}
The result of this step is shown in Fig.7.

\begin{figure}[8 h]
\begin{center}
    \includegraphics [scale=0.52] {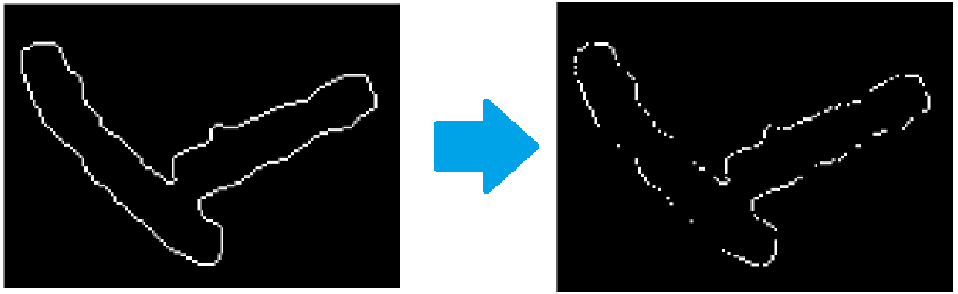}
 \end{center}
  \caption{Remaining pixels after applying VAMD criterion}
\end{figure}

\subsection{Sum of Distances among Total Points (SDTP)}

Because of noise in chromosomes images, after the aforementioned step, there may remain more than two points, so we should use another criterion to find the cross-points.
Let us assume after the previous step, M points have remained. We denote these points with $B_1$ to $B_M$ where M is the number of remaining points.
The cross-points are usually located in the middle of overlapping chromosomes (because overlapping chromosomes are formed by two or more chromosomes).
For each remaining pixel, we find the sum of distances between this pixel and other remaining pixels and we expect this sum in the cross-points to be less than the other points.
We can find the sum of distances from each pixel to other pixels as:
\begin{equation}
Dis(i) = \sum_{\stackrel{j=1}{j\neq i}}^M d( B_i, B_j)
\quad \text{for}\hspace{0.5cm} i=1:M
\end{equation}
where d($B_i, B_j$) is the Euclidean distance between i-th and j-th pixels on the boundary.

We can select two pixels with the minimum amount of $Dis(i)$ as cross-points.
However in some cases, this method can select the wrong points. For example, if one small chromosome touches a large chromosome by its end, this selection method will not work properly. Therefore, we have to use another criterion alongside this criterion in order to avoid such errors. 
In order to avoid mis-selection, we can use both VAMD and SDTP in our final decision criterion. Therefore, in cases that SDPT cannot select the right points, VAMD can help the algorithm avoid mis-selection. We defined a cost function which takes into account both VAMD and SDTP and selects two points with minimum amount of cost function as the cross-points.
\begin{equation}
\text{Cost}(i) = {Dis}(i) -  \lambda \times \Delta\theta_{i} 
\end{equation}

The parameter $\lambda$ should be a positive number that can control the effect of $\Delta\theta$ in the cost function. In order to minimize this function, one needs to minimize $Dis(i)$ and maximize $\Delta\theta_{i}$. The value of $\lambda$ can be determined by trial and error on a training set. Based on our simulation $\lambda$=1000 produces satisfactory results.

After applying this algorithm, we will choose two points with the two least values of Cost(i) as the cross-points and overlapping chromosome can be separated using the line between these two points. 
The result of this step is shown in Fig.8.

\begin{figure}[11 h]
\begin{center}
    \includegraphics [scale=0.52] {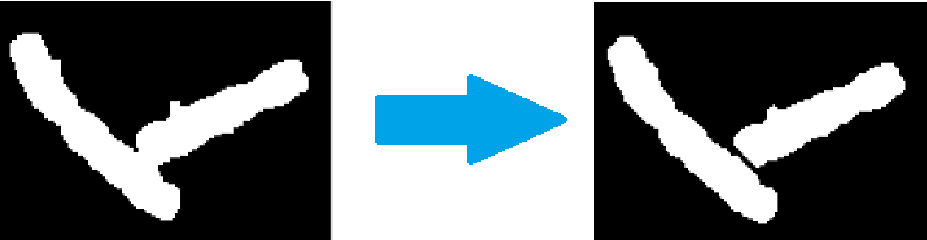}
 \end{center}
  \caption{Separation of  chromosome clusters with proposed method}
\end{figure}

\section{Separation of chromosome clusters with more than two chromosomes}
In the previous sections, we concentrated on chromosome clusters consisting of two chromosomes. In some cases, chromosome clusters may have more than two chromosomes. For binary images, it is difficult to separate a chromosome cluster with more than two single chromosomes in one step. However, we can easily separate this type of chromosome clusters by a multi-step algorithm. In fact, if a chromosome cluster consists of N single chromosomes, we can separate all single chromosomes in N-1 steps. We propose the following algorithm for separation of clusters with more than two chromosomes:
\begin{enumerate}
\item	Separate each chromosome cluster with previous methods. After separation we will have two new chromosomes.

\item For each new chromosome check whether it is a single chromosome or a chromosome cluster.

\item	If both of the new chromosomes are single, the algorithm is finished.

\item	If at least one of the new chromosome is a chromosome cluster, separate it using the procedure in the second phase.

\item	Continue this algorithm until all new chromosomes are single.
\\
\end{enumerate}

\section{Results}
By using the proposed algorithm, we analyzed 25 images containing a total number of 1150 chromosomes. There are about 62 touching or partially overlapping chromosomes in this data set.  %The number of chromosomes in our data set is not a lot. We could use a greater number of chromosome clusters in our data set(probably about 120 chromosome clusters), but most of them are similar to these 62 chromosome clusters and we want test our algorithm on a set of chromosome clusters that have different pattern.
We are interested in assessment of the ability of the proposed algorithm to successfully separate clusters into their composing chromosomes. 
We tested our algorithm on these 62 chromosome clusters and observed that this algorithm separates 57 chromosome clusters correctly. So an accuracy rate about 91.9\% is attained.
In Table I, we have reported the fraction of correct separations of touching and partially overlapping with respect to their total number. In this table, we also have reported a comparison with all similar results reported in the literature.
\begin{table}
\centering
\caption{proposed method accuracy, commpared to methods in the literature}
\begin{tabular}{|c|m{4cm}|c|}
\hline
Method & Number of 	touching or partially overlapping chromosoms & Accuracy \\
\hline
Ji (1989) [1] set 1 & \ \ \ \ \ \ \ \ \ \ \ \ \ \ \ \ \ \ \ $458$ & $95\%$ \\
\hline
Ji (1989) [1] set 2 & \ \ \ \ \ \ \ \ \ \ \ \ \ \ \ \ \ \ \ $565$ & $98\%$ \\
\hline
Lerner (1998) [4]  & \ \ \ \ \ \ \ \ \ \ \ \ \ \ \ \ \ \ \ $46$ & $82\%$ \\
\hline
%Shunren et al. 2003 [5] & \ \ \ \ \ \ \ \ \ \ \ \ \ \ \ \ \ \ \ $40$ & $92\%$ \\
%\hline
Grisan (2009) [9]  & \ \ \ \ \ \ \ \ \ \ \ \ \ \ \ \ \ \ \ $819$ & $90\%$ \\
\hline
Proposed method  & \ \ \ \ \ \ \ \ \ \ \ \ \ \ \ \ \ \ \ $62$ & $91.9\%$ \\
\hline
\end{tabular}
\end{table}

Fig.9 presents results of separation between six touching and partially overlapping chromosomes.
As we can see, this method provides very efficient separation of touching and partially overlapping chromosomes.

\begin{figure}[13 h]
\begin{center}
    \includegraphics [scale=0.6] {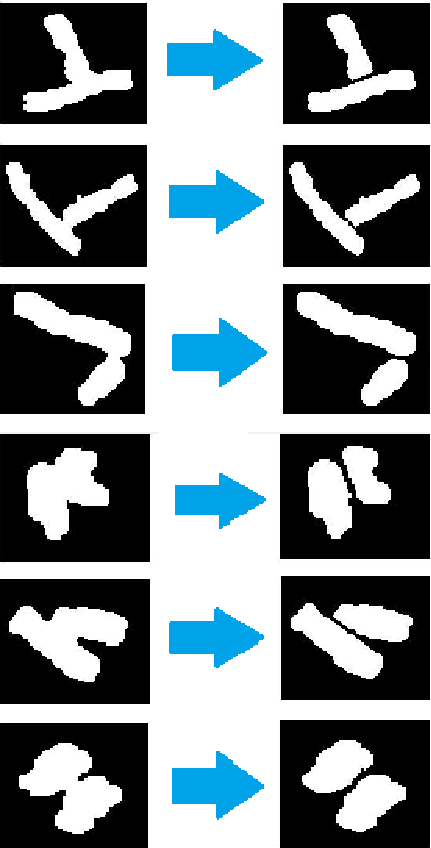}
 \end{center}
  \caption{Examples of separation of chromosome clusters using proposed method}
\end{figure}

\section{Conclusion}
In this paper, a geometric-based approach is proposed for chromosome segmentation. It uses three criteria for detection of chromosome clusters. After that, it uses a novel geometric method to find two cross-points on the boundary of clusters which can be used for extraction of cut-line. After the cut-line is found, we can decompose groups of chromosomes which touch and overlap each other. This algorithm is able to decompose clusters of touching or partially overlapping chromosomes that consist of more than two chromosomes. Another advantage of this method is that it can easily apply to any type of images, even binary chromosome images. In addition, due to use of geometric features of chromosomes which are independent of image type, the proposed scheme does not need multispectral images. 
\\ 
In future, we will focus on separation of completely overlapping chromosomes. For this purpose, first we should distinguish between touching chromosomes and overlapping chromosomes and apply the related algorithm to each class (separation algorithm for touching or partially overlapping chromosomes is different from separation algorithm for completely overlapping chromosomes). Separation algorithm for completely overlapping chromosomes is based on finding the cross section between two overlapping chromosomes and using it for separation of chromosome clusters.

\section*{Acknowledgments}
The authors would like to thank Prof. Hamid Aghajan for his invaluable help during this project. We would also thank Mr. Payam Delgosha, Mr. Amirali Abdolrashidi and Mr. Ali Hashemi for their useful comments on our project. 

\ifCLASSOPTIONcaptionsoff
  \newpage
\fi

\end{document}